# PRED18: Dataset and Further Experiments with DAVIS Event Camera in Predator-Prey Robot Chasing

Diederik Paul Moeys[1], Daniel Neil[1], Federico Corradi[1], Emmett Kerr[2], Philip Vance[2],
Gautham Das[2], Sonya A. Coleman[2], Thomas M. McGinnity[2], Dermot Kerr[2], Tobi Delbruck[1]
[1]Institute of Neuroinformatics, University of Zürich and ETH Zürich, Switzerland
[2]Intelligent Systems Research Centre, University of Ulster, Londonderry, Northern Ireland, UK

*Abstract*— Machine vision systems using convolutional neural networks (CNNs) for robotic applications are increasingly being developed. Conventional vision CNNs are driven by camera frames at constant sample rate, thus achieving a fixed latency and power consumption tradeoff. This paper describes further work on the first experiments of a closed-loop robotic system integrating a CNN together with a Dynamic and Active Pixel Vision Sensor (DAVIS) in a predator/prey scenario. The DAVIS, mounted on the predator Summit XL robot, produces frames at a fixed 15 Hz frame-rate and Dynamic Vision Sensor (DVS) histograms containing 5k ON and OFF events at a variable frame-rate ranging from 15-500 Hz depending on the robot speeds. In contrast to conventional frame-based systems, the latency and processing cost depends on the rate of change of the image. The CNN is trained offline on the 1.25h labeled dataset to recognize the position and size of the prey robot, in the field of view of the predator. During inference, combining the ten output classes of the CNN allows extracting the analog position vector of the prey relative to the predator with a mean 8.7% error in angular estimation. The system is compatible with conventional deep learning technology, but achieves a variable latency-power tradeoff that adapts automatically to the dynamics. Finally, investigations on the robustness of the algorithm, a human performance comparison and a deconvolution analysis are also explored.

## I. INTRODUCTION

CNNs [1] are taught from labeled data to extract useful features from their input example training set. They are widely applied in many scenarios such as: tracking [2][3], image- [4][5], face- [6][7] and location-recognition [8], image segmentation [9][10], self-driving cars [11][12] and so on.

Dynamic and Active Vision Sensors [13][14] combine Active Pixel Sensors' (**APS**) capabilities of generating intensity frames, like normal cameras, together with the Dynamic Vision Sensor (**DVS**) event output of [15]. Each pixel of the DAVIS sensor asynchronously produces events encoding for temporal contrast and concurrent intensity samples. Each pixel produces ON or OFF events whenever a positive or negative logarithmic change in brightness is detected. The sensor outputs the pixel location and time of the change, which is timestamped at microsecond resolution. The DVS output has a dynamic range of 120 dB and only produces data when the scene reflectance changes: i.e. in response to pixel brightness changes. The DAVIS data stream thus allows adaptive data-driven vision, where a combination of image frames and DVS event streams is used to drive vision at variable sample rate, depending on the scene dynamics.

In this work, we report the development of the open-source Predator Prey Dataset 2018 (**PRED18**), first introduced in [16][17], together with further experiments on it. This dataset, similarly to the ones of [18] and [19], was obtained with the DAVIS image sensor of [13], capturing both conventional frames and event-based data. The sensor was mounted one of two moving wheeled Summit XL robotic platforms following each other in a predator-prey fashion in a robot arena. Recordings were obtained under various environmental and control conditions. The Ground Truth (**GT**) of the prey position (but not size) in the Field Of View (**FOV**) was hand-labelled frame by frame and was used in the initial experiments of [16].

In [16][17], we reported a small CNN with only 4 outputs to process the event and frame data of the DAVIS sensor. The aim of the CNN was to give steering commands to the Summit XL predator robot within the arena, in order to chase another Summit XL prey robot. The network needed to detect the angle of the prey in the FOV of the on-board DAVIS sensor. In [16], inspired by the robot navigation of [20][21], this was achieved very similarly to the forest trail-following task of [22]: by separating the FOV into three parts and recognizing whether the prey was in the Left (**L**), Center (**C**) or Right (**R**) part of it. A fourth class, Non-visible (**N**), was used to indicate the lack of the prey in the scene. The four classes were sufficient to steer the predator efficiently after the prey.

In this work, we present further analysis and experiments on the PRED18 dataset. Firstly, we relabeled the dataset to include bounding box size. Secondly, we improved the algorithm to output an analog steering angle as well as relative analog predator-to-prey distance. This was done in order to smooth out the movements of the predator thanks to a finer prey position estimate and to slow it down inversely proportionally to the estimated distance. Thirdly, we studied a larger variety of CNN architectures, compared the accuracies with humans, increased inference efficiency by running the CNN in C on an NVIDIA Jetson TK1 embedded platform, obtained additional insights into the CNN operation by deconvolution analysis, and described the robot potential field control policy.

Sec. II introduces the robot setup, data collection and labeling, as well as DAVIS data preprocessing. Sec. III describes the optimization process behind the run-time CNN and it further develops the deconvolution analysis of [16], in order to visualize the receptive fields of the CNN. Sec. III.C also compares the accuracy of the CNN to human decisions. Sec. IV.A describes the C-based implementation which runs on the ROS controller and describes the robot control policy. The prey position vector extraction is also explained in Sec. IV.B. The predator behavior is described in Sec. V. Sec. VI then describes the final results. While in [16] the performance of the CNN was validated with closed-loop trial runs at the University of Ulster, in this work it assessed on the recordings of the trial runs of [16] because of the conclusion of the EU project VISUALISE and the availability of the testing arena.

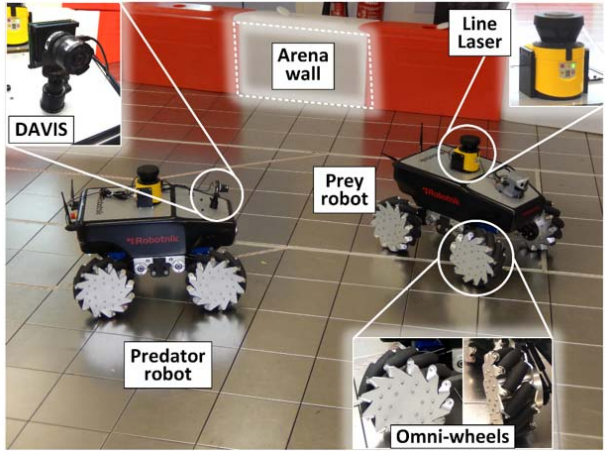

Fig. 1 Photo of the predator robot (left) and of the prey robot (right) in the robot arena of the University of Ulster. The predator robot holds the DAVIS sensor.

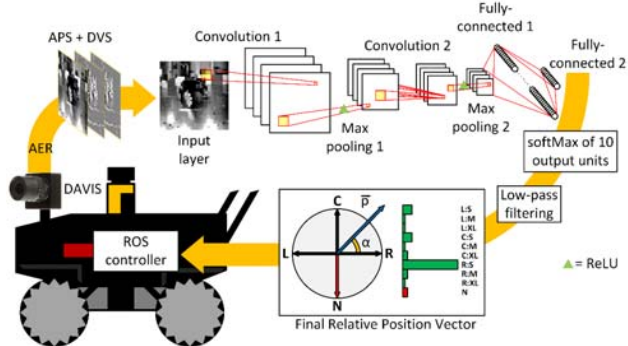

Fig. 2 Control system of the predator. A final relative angle and distance are computed from the position vector $\vec{p}$ as a function of the CNN's outputs.

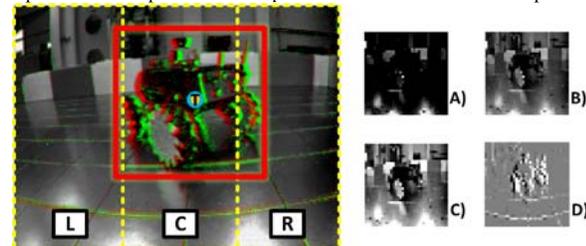

Fig. 3 Overlaying of APS and DVS data (red: OFF events, green: ON events). The FOV is divided in the 3 LCR regions and the prey GT location is labelled with a T. The prey size is labelled with a red square. A, B and C are the extracted 36x36 subsampled augmented APS frames. D is a subsampled DVS histogram.

## II. PRED18 DATA COLLECTION AND LABELING

The PRED18 dataset is available at [23].

### A. Robot and setup description

The predator/prey scenario was realized by two Robotnik Summit XL mobile robots (Fig. 1) [8] in a 9.5x6.7 m arena of the University of Ulster. The robots measure 750x540x370 mm and have omnidirectional Mecanum wheels with rollers on them, which allow skid-row kinematics. They weigh 40 kg and can reach up to 3 m/s (4 body-lengths/s). Due to damaging crashes in the trial runs, their speed was limited to 2 m/s.

The robots are equipped with an Inertial Measurement Unit (**IMU**) and a line laser scanner to detect objects and avoid collisions. The Robot Operating System (**ROS**) framework [9], running on the embedded Linux PC with Intel Core i7 processor, receives commands and controls the robot. Communication to a base station, to observe the status of ROS, is possible thanks to an embedded Wi-Fi 802.11n module. The predator robot is fitted with the DAVIS camera, which uses a short 2.6 mm wide-angle lens that provides a horizontal field of view of 81°. Various lighting conditions were created in the arena by turning the room lights on and off and shutting the blinds. Weather conditions also affected lighting through the unshaded windows. Some recordings include the flicker on the floor of the Vicon tracking system (turned off afterwards). Finally, the floor of the arena, covered in stripe patterns from other experiments, was made of reflecting conductive material. Background objects with strong and weak patterns also appear above the arena walls.

### B. Data collection and pre-processing

PRED18 consists of twenty recordings, made up of DVS events and APS frames, for a total of 1.25h. These were obtained from the DAVIS sensor mounted on the predator robot roaming around the arena together with the prey robot and, sometimes, walking humans. The prey was always controlled by a human. Example data is shown in Fig. 3. In the first ten recordings, prey and predator are human-controlled, while in the last ten recordings, the predator was controlled by the *LCRN* plus potential field control described in [16].

The 240x180 APS frames of the DAVIS were recorded at 15 fps and subsampled to square images of size 36x36, 54x54 and 72x72. The reason for subsampling was to find the lowest image resolution possible allowing still a good classification accuracy and at the same time a small, and therefore fast, CNN. The image dimensions are multiples of 3 to ensure the divisibility in the 3 steering regions. Two datasets were composed: one was kept raw and one was filtered for uncorrelated noise activity with a background activity filter, not used in [16]. This algorithm in jAER [24], the Java-based platform processing the DAVIS data, filters out the uncorrelated noise due to the parasitic photocurrent in the reset switch [15] of the DVS pixel by setting spatial and temporal correlation limits. Events that do not occur within a certain distance of the previous event and within a certain time (10ms) are ignored. The DVS datasets were then integrated into 2D histograms of 5k events with same dimensions of the APS frames. Histograms of DVS data allow compatibility with state-of-the-art CNN technology, while taking advantage of the activity-dependent frame rate of the DVS. The subsampling was obtained by truncated division of the event coordinate. Starting from a grey value of 0.5 out of 1, each event integrated contributes to a fractional increase (ON) or decrease (OFF) in gray values, with clipping at +/-1 to handle hot pixels with high firing rates. DVS histograms are then normalized at 3 standard deviations of their value distribution from the original histogram mean. This was done to further remove outlier pixel and increase the contrast of meaningful features. The 0.5 gray level of a DVS histogram corresponds to zero events. DVS frame normalization is arranged to preserve the 0.5 value to avoid unwanted gray level flickering dependent on the ratio of ON and OFF events.

APS data was augmented by over- and under-exposing APS data by shifting the gray values of the frames by a fixed amount and clipping out-of-range values. The data was also mirrored horizontally to even out the number of *L* and *R* examples. Such augmentation, resulted in a total of 500k images (a shuffled mix composed of 45% APS frames and 55% DVS histograms).

*C. Data labeling*

The Ground Truth (GT) location and size of the prey robot in the FOV were obtained by manual labelling of the robot position in jAER. The mouse pointer indicated the position while the mouse wheel controlled a bounding box for the size. Labeling used the jAER filter *TargetLabeler*. Using the analog position, the frame was labeled *L*, *C*, or *R* depending on the vertical third of the image (27° of the 81° FOV) that the prey was centered in. If the prey was not visible, then the frame was labeled *N*. The size of the prey was labeled *S*, *M* or *XL*. Data was easily labeled within two passes (one for prey position, one for size) through each recording at about half of real time speed.

Fig. 4 shows the prey size distribution in all recordings. Two size thresholds, $thr_L$ and $thr_H$, were set to match one standard deviation of the prey size with respect to its mean. Each image of the dataset would therefore have one of the 9 possible combinations of *L*, *C*, *R* and *S*, *M*, *XL*, with *N* as the tenth category. Examples of labeled inputs are shown in Fig. 5. 80% of each recording of the dataset was used as the training image dataset. The remaining 20% was used as the test dataset. The video frames were not shuffled before their division into training and test sets to avoid having similar consecutive frames in both sets to prevent overfitting. The dataset was imported in .lmdb format to the Caffe deep learning framework [25] for training.

### III. CNN STRUCTURES AND ANALYSIS

This section discusses the architectural explorations to determine an optimal CNN for the steering task, insights about its operation, and comparisons of accuracy with humans.

*A. Optimization process*

The input size was changed to 36x36, 54x54 and 72x72 to evaluate the effect on accuracy for the detection of the prey when far away. 36x36 was the smallest input for which the prey could still be identified when far away by human inspection.

Using the same notation used in [16], the final selected network has the architecture $10C5\text{-}R\text{-}2S\text{-}20C5\text{-}R\text{-}2S\text{-}100F\text{-}R\text{-}10F$. The first convolutional layer has $n = 10$ output feature maps with 5x5 kernels (denoted $nC5$), with ReLU activation function $(R)$ followed by a 2x2 max pooling subsampling layer $(2S)$. Unlike in [16], the convolution uses zero padding (2 per side of the input layer) in order not to lose resolution and information on the edges of the input. Another convolution layer with $m = 20$ output feature maps with 5x5 kernels ($mC5$) with zero-padding and ReLU activation function then follows together with a max pooling layer with, again, stride 2. Following the two convolution and subsampling layers is a fully connected layer of $q = 100$ neurons. This layer, followed by ReLU activation, is crucial for classification as it expands the dimensionality of the preceding convolutional layer. The final classification layer ($10F$) consists of 10 fully connected output neurons to which the softmax function is applied. These represent *L:S*, *L:M*, *L:XL*, *C:S*, *C:M*, *C:XL*, *R:S*, *R:M*, *R:XL*, *N*.

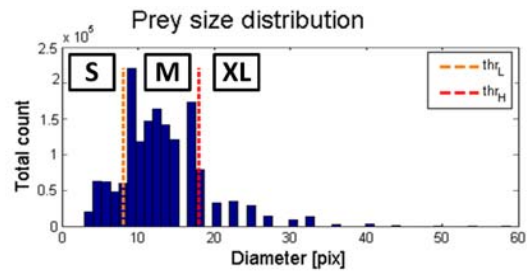

Fig. 4 Prey size distribution for a 36x36 input, illustrating size discrimination.

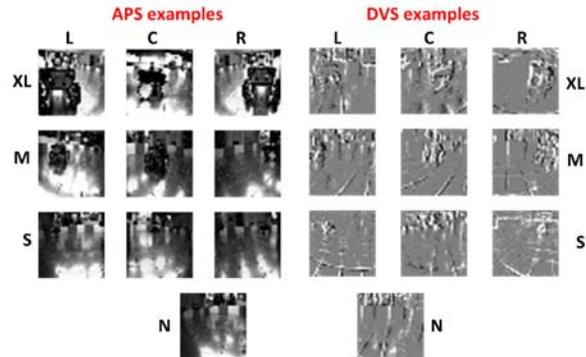

Fig. 5 Example 36x36 labeled APS frames and DVS histograms for all 10 categories consisting of *N* and all combinations of *L*, *C*, *R* and *S*, *M*, *XL*.

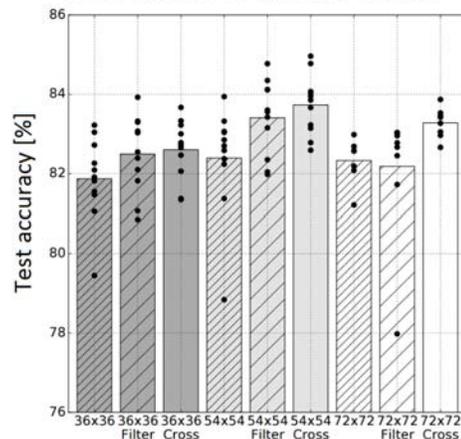

Fig. 6 CNN test accuracies (for various hyper-parameters) for 3 input sizes and 3 dataset types. Bars represent the mean of the single network results (black dots distributions). The first accuracy of the 3 sets indicates that the nets have been trained and tested on the non-background-filtered dataset. *Filter* indicates nets trained and tested on the background-filtered dataset. *Cross* indicates nets trained on non-background-filtered data but tested on background-filtered data. A noisier training dataset improves the network robustness.

The CNN was selected through manual optimization, testing a fixed set of networks for the 3 input-sizes (36, 54 and 72). The idea was to maximize accuracy while minimizing network size and thus real-time latency. Some results of this optimization on different datasets are visible in the bar chart of Fig. 6. Overall, for a 36x36 input image, the accuracies of the CNNs saturate at

~83-84%. The limitation in accuracy lies in the ambiguity of the hand-labelled data. When the prey robot moves across *LCRN* boundaries, the quantized label is ambiguous. It can be seen in Fig. 7, where the errors in estimating the position of the prey rise when the x-position label approaches the boundaries of the *LCRN* regions and human judgement of position plays a significant role together with quantization. Although the dataset is augmented by mirroring, perfect symmetry does not appear in the histogram of Fig. 7, because the quantization is done with a flooring operation. The same boundary problem happens for size labels but is not very visible in the right histogram of Fig. 7 because the effect of the *LCRN* boundaries hides this and as the majority of all images (68.2% by definition) have *M* size.

Fig. 6 also illustrates the effect of input size on accuracy while all other hyper-parameters are kept the same. The slight increase in accuracy of 1% from 36x36 to 54x54 input size, corresponds to an improvement for images labelled S where the prey robot is only a few pixels wide. Improvements also appear for sizes where quantization is critical (at the boundaries of sizes). This proves that the network is mainly limited by the dataset used and only for a small percentage of images by the resolution of the input. For the 72x72 input, the accuracy starts to decrease again by about 0.5%, probably because kernels are too small. All other hyper-parameters were kept constant to minimize the size of the network and a possible over-fitting.

Finally, in Fig. 6, it also possible to see that nets indicated as "36x36/54x54/72x72 Cross" (CNNs trained on non-background-filtered data, but tested on background-filtered data) have a higher accuracy than nets trained and tested on the same dataset (either "36x36/54x54/72x72" non-filtered or "36x36/54x54/72x72 Filter", where the background filter was used). Training on a noisier dataset makes the CNN more robust. Our attempts to improve the network's accuracy by using multiple frames at different time intervals, did not improve accuracy by more than 0.5-1%, due to the noisy labels limitation.

For these reasons, the 36x36 input $10C5$-$R$-$2S$-$20C5$-$R$-$2S$-$100F$-$R$-$10F$ (the peak-performing net of 36x36 Cross of Fig. 6) was selected as the runtime network. The normalized confusion matrices of the chosen CNN tested on 36x36 and 54x54 input images are shown in Fig. 8. As the input resolution increases, the ambiguity in size estimation decreases (for example *C:M* and *C:XL*) and the *S* size accuracy increases.

### B. Deconvolution analysis

In Fig. 9, the guided-backpropagation method of [26] to visualize saliency was used to understand which part of the network's input image causes the greatest absolute change in the output of several of the layer 100F fully-connected neuron units just prior to the classification layer. For this example, the prey robot is located in the upper left corner of the image. As can be seen in Fig. 9, individual neurons' receptive fields are highly localized to specific regions of the image because the network was trained to spatially localize the prey. Certain neurons demonstrate a strong response to the wheels of the robot, while others are more sensitive to the bright windows and reflections on the floor, which can hint at the presence or absence of the robot from a portion of FOV. The low input resolution, however, does not allow the identification of highly detailed patterns such as the ones in [27]; making use of 72x72 input images to perform the same analysis does not reveal further insights. The network's depth and dataset are tiny compared to the size of CNNs trained on ImageNet [27], and the subject is not cleanly localized in the center of each example, making visualization of features less consistent.

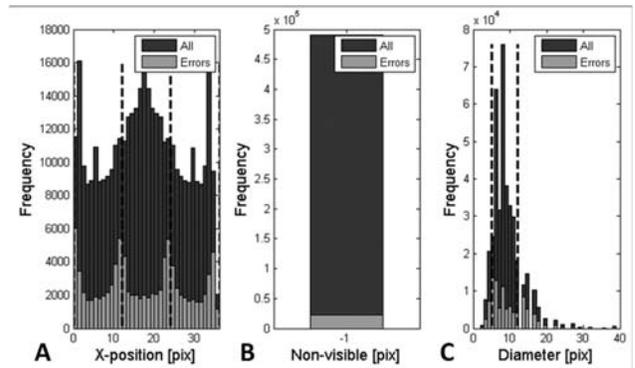

Fig. 7 **A:** histogram of x-position of the prey robot in the dataset compared with the errors of the CNN on such frames. **B:** histogram of N compared with the errors of the CNN. **C:** histogram of all sizes of the prey robot in the dataset compared with the size errors. As can be seen, at least for position, larger errors occur at *LCRN* regions boundaries (dotted lines at x positions 0, 12, 24, 36).

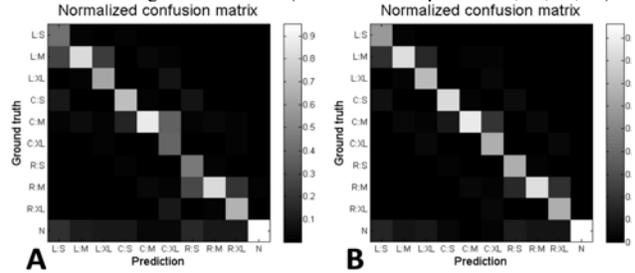

Fig. 8 Normalized confusion matrices of the 10C5-R-2S-20C5-R-2S-100F-R-10F network trained on the unfiltered 36x36 (A) and 54x54 (B) datasets.

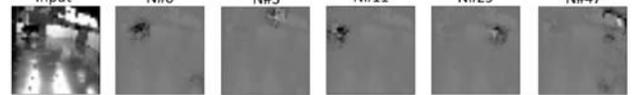

Fig. 9 Saliency map visualization: input image and resulting images weighted in transparency depending on the absolute effect of input pixels on the output of some of the 100F fully-connected layer units.

### C. Human performance comparison

Since the accuracy of the CNN is lower than the commonly reported for state of the art CNNs on MNIST, a Matlab program was developed to compare it with human performance. The Matlab Graphic User Interface (GUI) of Fig. 10 shows a randomly sampled image to the human test subject. The image is marked with the *L*, *C* and *R* regions separated to let the user decide by pressing 'a', 's', 'd' or 'n' whether the prey robot is in *L*, *C*, *R*, or *N*. The user then estimates the size of the prey robot by pressing '1', '2' or '3'. To estimate the size, one compares the size of the prey with the size of the squares drawn alongside in blue and red (representing $thr_L$ and $thr_H$). Nine test subjects were tested twice for 25 images. Although if the images were presented as a camera stream they would be more understandable by a human, these were presented individually in a random order, for a fair comparison with the CNN, which does not exploit the temporal sequence of images. Their accuracy is shown in Fig. 11 versus the time it took them to decide. For untrained subjects, who were only shown part of the original non-subsampled recording to recognize the robot, the accuracy is 52% on average, with a mean 8.1s response time. In comparison, a trained subject (author DPM) can reach an average accuracy of 76% and 2.5s response time. This performance can be compared to the 2ms feed-forward latency and 83% accuracy of the chosen CNN.

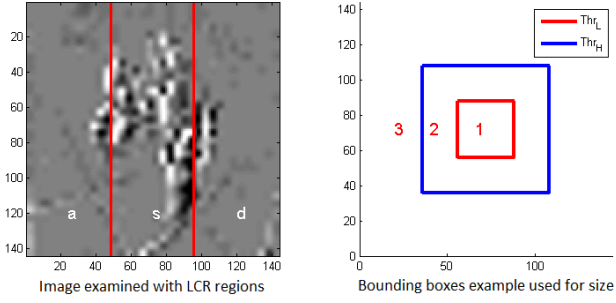

Fig. 10 Matlab interface for human performance comparison showing the image under examination delimited by the *L*, *C* and *R* regions and the two bounding boxes corresponding to $thr_L$ and $thr_H$ used to set *S*, *M* and *XL*.

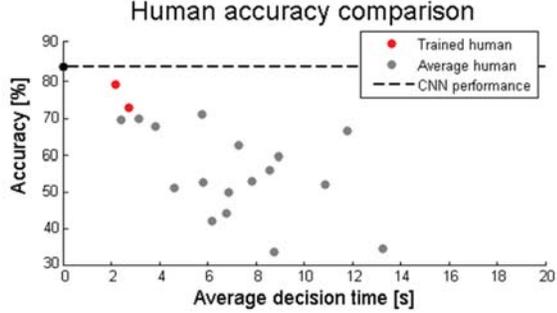

Fig. 11 Human performance of two trial runs of 9 subjects with an average of 25 images each compared to the CNN performance, plotted vs decision time.

## IV. CNN JAVA AND C IMPLEMENTATIONS

The selected CNN has a total of 5.2k CNN weights and 162k fully-connected layer weights. It requires a total computation of about 2MOp per frame. APS and DVS frames are merged in the same stream and processed, as soon as available, by the same runtime network. The CNN was first implemented in Java (filter *eu.visualize.ini.VisualiseSteeringConvNet*, in jAER [24]) where DVS histograms and APS frames can be generated and pre-processed. The same filter implements the chosen CNN, which computes the ten output classes and whose weights are loaded via an .xml file. This .xml file is generated from the conversion of the Caffe files containing the network weights thanks to the script *cnn_to_xml.py*, also available at [24] under *scripts/*. The CNN Java-based engine also allows synchronizing the GT of the recording, via *eu.visualize.ini.TargetLabeler*, to validate the accuracy of the network. The loaded weights and the intermediate processing of every layer can also be visualized real-time. A feed-forward computation, for the CNN of [16] required 4 ms on the predator's Intel i7 processor with Java JDK1.8, which is sufficient to run in at up to 250 fps. The processing delay of the $10C5$-$R$-$2S$-$20C5$-$R$-$2S$-$100F$-$R$-$10F$, on the same computer takes 6 ms (167 fps). The runtime CNN output is shown in Fig. 12 for a particular input image.

### A. C implementation

To simplify the number of steps and exploit the efficiency of Caffe, the system was also implemented in C/C++ code running on the embedded NVIDIA Jetson TK1 developer platform in CPU mode. The DAVIS sensor was connected to the TK1 using the USB2.0 host controller port. With 36x36 gray scale input images, the computation of a single input image on CPU takes on average <2 ms, meaning that the system is able to process at ~500 fps. Using the GPU on the TK1 does not improve the result by more than 0.1 ms, since most of the delay is due to image loading and since the CNN is small and it is not processed in batch mode, where GPU acceleration is effective. The code is open-sourced[1].

### B. Relative analog position extraction

We chose to train an angle and size classifier by extending our previous work. Using the classifier output, an analog regression of angle and a distance can be computed from the 10 output units of the CNN. The result is merged in a position vector $\vec{p}$ of the prey relative to the predator (Fig. 2), with angle $\alpha$ and modulus $|\vec{p}|$ equal to the distance.

Angle $\alpha$ is computed from the x and y projections $dX$ and $dY$ obtained from the softmaxed output units $o$. $dX$ is

$$dX = \frac{o(R:S)+o(R:M)+o(R:XL)}{3} - \frac{o(L:S)+o(L:M)+o(L:XL)}{3} \quad (1)$$

$dY$, the vertical component, is computed in (3); $o(N)$ represents, in this case, "behind".

$$dY = \frac{o(C:S)+o(C:M)+o(C:XL)}{3} - \frac{o(N)}{r} \quad (2)$$

where $r$ is a scaling parameter. The angle $0°<\alpha<180°$ is computed by

$$\alpha[°] = \begin{cases} \frac{-180°}{\pi} atan\left(\frac{dX}{dY}\right) + 90° & if\, dY > 0 \\ \frac{-180°}{\pi} atan\left(\frac{dX}{dY}\right) + 270° & if\, dY < 0 \end{cases} \quad (3)$$

The magnitude $|\vec{p}|$ of the $\vec{p}$ is first computed from

$$s(S) = \frac{o(L:S)+o(C:S)+o(R:S)}{3}$$
$$s(M) = \frac{o(L:M)+o(C:M)+o(R:M)}{3} \quad (4)$$
$$s(XL) = \frac{o(L:XL)+o(C:XL)+o(R:XL)}{3}$$

The modulus $|\vec{p}|$ is then computed from

$$|\vec{p}| = \frac{(s(S)+s(M)/2+s(XL)/3)}{\kappa} \quad (5)$$

where $\kappa$ is a scaling parameter. $|\vec{p}|$ is only computed if $o(N)$ is not the largest output probability of the CNN. The weights multiplying the sizes in equation (6) were chosen to have a simple distance estimation varying linearly with object size.

Finally, the obtained $|\vec{p}|$ and $\alpha$ are low-pass filtered since they would otherwise change too rapidly. Just like in [16], the CNN output can still be digitized: the maximum of the units containing *L*, *C*, *R* or *N* can be selected as the rough location and similarly for the maximum of the units containing *S*, *M* or *XL*. Logical constraints can then be applied like to validate the CNN prediction. With the same location constraints of [16] (it is impossible to go from $L$ to $R$ or $C$ to $N$), the size constraint of the impossibility to switch between $S$ and $XL$ is added.

---

[1] https://github.com/inilabs/caer/tree/7a063137d739503c62feae2267ff1b2326208431/modules

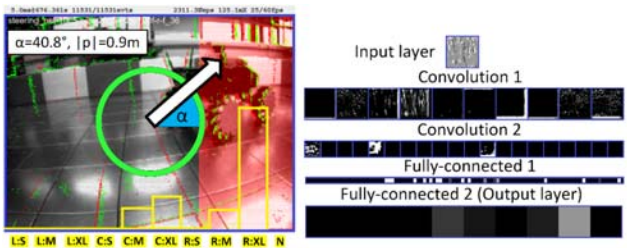

Fig. 12 JAER output visualization of the processing of the chosen runtime network. Left: digital final network decision (red), network outputs (yellow bar charts) and arrow (white) representing the relative analog position vector. $|\vec{p}|$ and $\alpha$ are estimated in realtime. Right: input and output feature maps of the convolution and fully-connected layers for the current input.

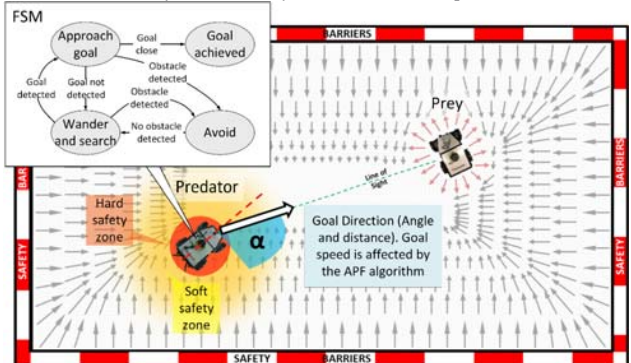

Fig. 13 Robots in the arena and repulsive fields surrounding obstacles. The predator identifies the position of the prey by angle and distance and plans its route while avoiding obstacles. The FSM of the predator is also shown.

## V. ROBOT BEHAVIOR

In [16], the *LCRN* position was sent out together with a sequence number, to a ROS module controlling the robot movements over a local UDP socket. This caused the loss of data packets whenever too many decisions were sent out in a row from jAER to ROS. In the new C implementation, UDP is no longer needed as cAER can call ROS directly. This makes it easy to send as many positions and angles as possible to the ROS controller. $\alpha$ can be used to scale the angular velocity of the predator and $|\vec{p}|$ distance to set the forward speed. $\alpha$ and $|\vec{p}|$ can be quantized to 20° and 1 m steps respectively to send out commands to ROS only when the change is significant.

### A. Finite State Machine

A Finite State Machine (FSM) control algorithm for a robot avoiding obstacles while chasing a dynamic goal was developed (Fig. 13). The FSM states are described as follows.

1) *Avoid*: The predator uses the potential fields-based obstacle avoidance algorithm outlined in Sec. V.B to avoid the detected obstacles. This state has priority at all times.
2) *Wander and Search*: When there is no detected obstacle in the robot's path (*N*), the predator explores the area while continuously searching for the prey. If the predator detects its goal, it then moves to the *Approach Goal* state.
3) *Approach Goal*: When the predator identifies its prey within its environment, it begins to approach it with the aim of positioning the prey in the center of its FOV (*C*, about 27° wide). When $L$ or $R$ decisions are received from the CNN, the predator's motion is composed of a rotation with an angular velocity of $\pi/3$ rad/s and of the maximal allowed forward linear speed. The latter varies between 1.5-2 m/s to avoid crashes in the small robotic arena. Speeds higher than 2 m/s make impossible for the prey's teleoperator to evade the predator. If the predator detects an obstacle at any stage during its movement towards the goal, it slows down and it switches to the *Avoid* state until it is at safe distance from the obstacle. It then begins to search for its goal again and approaches it upon identification. If the prey suddenly becomes non-visible, the predator just spins in place at $\pi/2$ rad/s in the last direction it saw it (either $L$ or $R$).
4) *Goal Achieved*: The goal state is reached when the prey robot is detected by the C unit and the distance sensor signals a small distance. After a timeout period to allow the prey to escape, the state is switched back to wander and search.

### B. Modified potential field algorithm for obstacle avoidance

There were no obstacle avoidance algorithms available within ROS that did not require an accurate goal location or path planning. Therefore, we developed an obstacle avoidance system based on the modified Artificial Potential Field (**APF**) [28] algorithm for mobile robots, which works within the FSM-based control algorithm. The APF approach considers that obstacles within the environment are "surrounded" with repulsive fields. These fields increase in strength the closer they are to an object. The code is open-source from *github.com/uu-isrc-robotics*. Force vectors are computed from laser range readings obtained from the on-board laser sensor. These vectors are used to compute repulsive fields. Any physical item which can be detected by the robot's laser range finder (arena barriers and the prey) is considered an object and thus will have repulsive fields associated with it (Fig. 13). The closer the robot is to the object the stronger the repulsive field.

The obstacle avoidance scheme has been implemented with two concentric hard (orange) and soft (yellow) Safety Zones (**SZ**), portrayed in Fig. 13 as the circles around the predator robot. The hard SZ, with highest priority, is set using a radius of 0.7 m from the centre of the robot (the turning radius plus a safety margin). Any object within 0.7 m of the robot center will make the robot immediately stop and attempt to rotate to a direction which corresponds to the path with least repulsive force, estimated over the whole potential field, to avoid any damage through collisions. The soft SZ is set with a radius of 1.5 m from the center of the predator, thus, any objects within this radius will trigger a reduction of linear and rotational velocities proportional to the strength of the potential field. This results in the robot avoiding any objects with a smooth motion. The obstacle avoidance behaviour always takes priority over all other behaviours. Unlike the predator-prey approaches in the literature (e.g. [43][44][31][32]), our approach implements a dynamic goal, i.e., the goal does not have to be specified before the algorithm commences, but can change during runtime.

## VI. RESULTS AND DISCUSSION

### A. Analog position estimation

The performance of the previous digital *LCRN* classification algorithm of [16] was directly tested on 8 different trial runs (one visible at [17]). The validation of the analog output $\alpha$ and $|\vec{p}|$ was estimated on the recorded data, for 36x36 input. Fig. 14A shows the correlation of $\alpha$ with the GT angle known from the labeled target position. As can be seen, the strong correlation is 0.9 with a 1.09 linear relationship. The plot can only show the correlation when for a visible prey. In **Fig. 14A**, $\alpha$ is simply rescaled between -40.5° and 40.5° because the real FOV of the DAVIS sensor (as well as the GT) is 81°. Overall, for all trial runs of [16], the mean absolute difference between angular GT and $\alpha$ is ~7.1°/81°=8.7%. Error jumps between opposite sides of the FOV (for example, -40° to suddenly 40°) do not appear in the graph and do not affect the percentage error because they are filtered out by the logical constraints mentioned in IV.B. Part of the error is due to noisy human hand-labeling. This value is below the quantization error of 27°/2=13.5° (17%) of the digital estimation. This makes the analog estimation better than the digital one, although the nonlinearity of the softmax output is clearly visible in the stepped nature of the scatter plot **Fig. 14A**.

For $|\vec{p}|$, there was no GT. $|\vec{p}|$ is therefore plotted versus the inverse of the GT size. As can be seen in **Fig. 14B**, the correlation is 0.66 but it is much stronger for prey sizes larger than 10 pixels wide (0.1< on the x axis). For prey sizes <10 pixels wide, the prey is often detected as *N*, because of the great strength of $o(N)$, due to the biased dataset (consisting of ~50% *N* data). This problem can be either mitigated by increasing the factor $r$ dividing $o(N)$ in equation (2) at the cost of more noise or by increasing the input resolution at the cost of more latency. Interestingly, although the prey's y position does not vary by more than 10 out of 36 pixels in the FOV, a correlation identical to the one of **Fig. 14B** was observed between $|\vec{p}|$ and y position. A video of the real-time results in jAER is available at [33].

### B. Data-driven computation

The overall runtime CNN accuracy on the test set, comprising the mixture of APS and DVS is 83%: 79% on DVS data and 87% on APS. Training the network on just APS data resulted in slight overfitting. The addition of DVS training data improved the generalization of the network. The difference in accuracies is due to two factors: the noisier DVS histograms and the DVS histograms where both predator and prey are still. Indeed, in the case of third-party motion (a person walking by in the FOV), the 5k events of the DVS histogram are filled up with events unrelated to the prey, making detection impossible.

The advantage of the use of the DAVIS sensor presented in this predator/prey application is that computation is data-driven. More DVS frames to be processed are produced when more activity is detected, i.e. when objects move quicker in the FOV. It means computation is automatically modulated by activity. The APS frame rate and DVS histograms rate distributions can be seen in **Fig. 15**. The mean APS frame rate is 17 Hz, but the DVS histogram rate spans 0.02-1580 Hz. The DVS frame rate, which is fed to the network as soon as a DVS histogram is integrated, is limited at 500 Hz, due to the 2 ms feedforward pass processing delay. Extra DVS frames are dropped if the time interval is less than 2 ms. If an APS camera would need to match the highest DVS histogram rate possible with the current system, it would need to sample at 500 Hz, a factor of 12X more computations. Increasing the APS frame higher than 17 Hz is also not possible since it is limited by the minimum exposure duration.

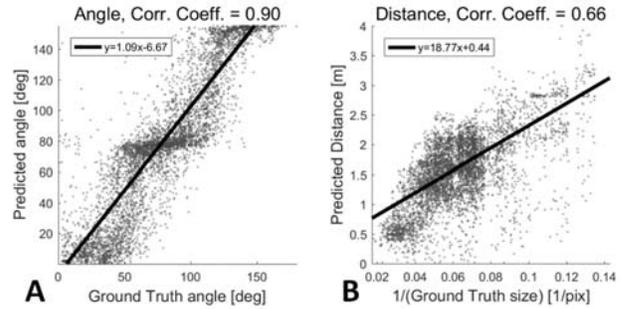

Fig. 14 Correlation plots of estimated (**A**) angle (0°<$\alpha$<180°) and (**B**) position.

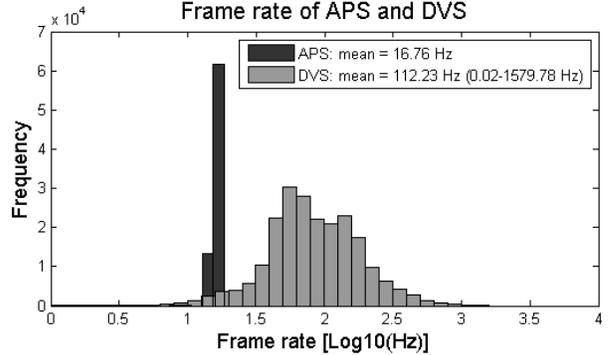

Fig. 15 DVS histogram rate and APS frame rate distributions.

In case of no ego-motion and no motion in the scene, the DVS should be turned off. This happens automatically if the background activity filter is active. The 5k events needed to fill one histogram do not fill up with these background noise events and the DVS does not produce any output. Using the background activity filter in normal ego-motion allows a clearer output as only strong contours appear. This makes it easier to discern the robot when it is far away in the scene and the few pixels composing the prey are precious.

While the DVS part of the DAVIS already produces a variable frame-rate, the frame-rate of the APS can be adjusted according to the DVS activity detected when no ego-motion is present. For a DVS event rate lower than 1keps, the APS frames could be turned off. This saves power and computation time as nothing substantial has changed in the scene. In the opposite case, the frame rate can be increased to 10 Hz. For a higher number of events per second but no ego-motion, the APS is still kept active as the object moving in front of the sensor might be a person passing by (hiding the robot in the DVS histogram).

## VII. CONCLUSION

This paper described the collection, labelling and processing of the open-sourced PRED18 predator-prey dataset. Initial results and research on the application of conventional deep learning technologies together with the silicon retina in a robotic scenario were further investigated and compared. Exploiting the advantages of both technologies, namely accuracy and data-driven capabilities, a closed-loop system was initially designed in [16] and further improved in this work.

The optimization process to reduce the size of the CNN and the number of operations it has to compute was paired with a careful reduction in input data by choosing a behavior for the sensor and by further removing uncorrelated activity. Nonetheless, such a small network could efficiently extract angular and size estimates of the target with a linear combination of its output probabilities. As discovered in the closed-loop runs, the 83% first estimate of accuracy is only dependent on labeling and still outperforms human performance on this difficult task (average accuracy of 52% and mean 8.1s decision time). The exact accuracy number is also irrelevant when post-processing can be applied and an analog angle can be extracted with mean 8.7% error in analog estimation. The large amount of augmented data, also makes up for the presence of human error in the quantized labeling.

Runtime was significantly decreased down to 2 ms per forward pass by the speed-up due to the incorporation of the Caffe machine learning software into the new C-based version of the network and due to the use of a dedicated NVIDIA Jetson TK1 Embedded platform's CPU.

The PRED18 dataset, runtime CNN and C/Java-implementations are open-sourced and available at [35][24].

ACKNOWLEDGEMENTS

This research is supported by the European Commission project VISUALISE (FP7-ICT-600954), SeeBetter (FP7-ICT-270324) and Samsung. We would like to thank the Sensors group at INI Zürich, Luca Longinotti from iniLabs GmbH, and the Intelligent Systems Research Centre of the University of Ulster.